%% file: main.tex
\documentclass[11pt,a4paper]{article}
\usepackage[hyperref]{naaclhlt2018}
\usepackage[utf8]{inputenc}
\usepackage{times}
\usepackage{url}
\usepackage{latexsym}
\usepackage{graphicx}
\usepackage{amsmath}
\usepackage{enumitem}
\usepackage{amsfonts}
\usepackage{boxedminipage}
\usepackage{color}
\usepackage{bm}
\usepackage{amssymb}
\usepackage{times}
\usepackage{bbold}
\usepackage{array}
\usepackage{multirow}
\usepackage{appendix}
\usepackage{sidecap}
\usepackage{mathtools}
\usepackage{todonotes}
\usepackage{xcolor}
\usepackage{comment}

\usepackage{caption}

\DeclareCaptionFont{10pt}{\fontsize{10pt}{12pt}\selectfont}
\newcommand{\ignore}[1]{}

\newcommand{\cut}[1]{}

\newcommand{\norm}[1]{\lVert#1\rVert}
\title{
Discourse-Aware Neural Rewards for Coherent Text Generation
}

\author{Antoine Bosselut$^{1}\thanks{\hspace{2mm}Work done while author was at Microsoft Research}$, Asli Celikyilmaz$^2$, Xiaodong He$^3$, \\
\textbf{Jianfeng Gao}$^2$, \textbf{Po-Sen Huang}$^2$ \and \textbf{Yejin Choi}$^{1,4}$\\
$^1$Paul G. Allen School of Computer Science \& Engineering, University of Washington \\
$^2$Microsoft Research \\
$^3$JD AI Research \\
$^4$Allen Institute for Artificial Intelligence\\
\texttt{\{antoineb,yejin\}@cs.washington.edu} \: \texttt{\{xiaodong.he\}@jd.com} \\
\texttt{\{aslicel,jfgao,pshuang\}@microsoft.com}
}
\begin{document}
\aclfinalcopy
\maketitle
\begin{abstract}

In this paper, we investigate the use of discourse-aware rewards with reinforcement learning to guide a model to generate long, coherent text. In particular, we propose to learn neural rewards to model cross-sentence ordering as a means to approximate desired discourse structure. Empirical results demonstrate that a generator trained with the learned reward produces more coherent and less repetitive text than models trained with cross-entropy or with reinforcement learning with commonly used scores as rewards.

\end{abstract}
\input{intro}

\input{reward}
\input{agent}
\input{reinforce}
\input{setup}
\input{experiments}

\input{related}
\input{conclusion}
\bibliography{reciperl}
\bibliographystyle{acl_natbib}
\include{appendix}
\end{document}

%% file: intro.tex
\section{Introduction}
\label{sec:intro}
Defining an ideal loss for training text generation models remains an open research question. 
Many existing approaches based on variants of recurrent neural networks \cite{lstm,cho2014learning} are trained using cross-entropy loss  \cite{bahdanau2014neural,NIC,ImgAttn,deepsummrush}, often augmented with additional terms for topic coverage or task-specific supervision \cite{checklist,reflm}.

Training with cross-entropy, however, does not always correlate well with achieving high scores on commonly used evaluation measures such as ROUGE \cite{rouge}, BLEU \cite{bleu}, or CIDEr \cite{cider}. Another current line of research therefore explores training generation models that directly optimize the target evaluation measure \cite{googlemt,mixer,rlsummsocher,scic} 
using reinforcement learning methods such as the REINFORCE algorithm \cite{reinforce}. 

\begin{figure}[t!]
\includegraphics[width=0.48\textwidth]{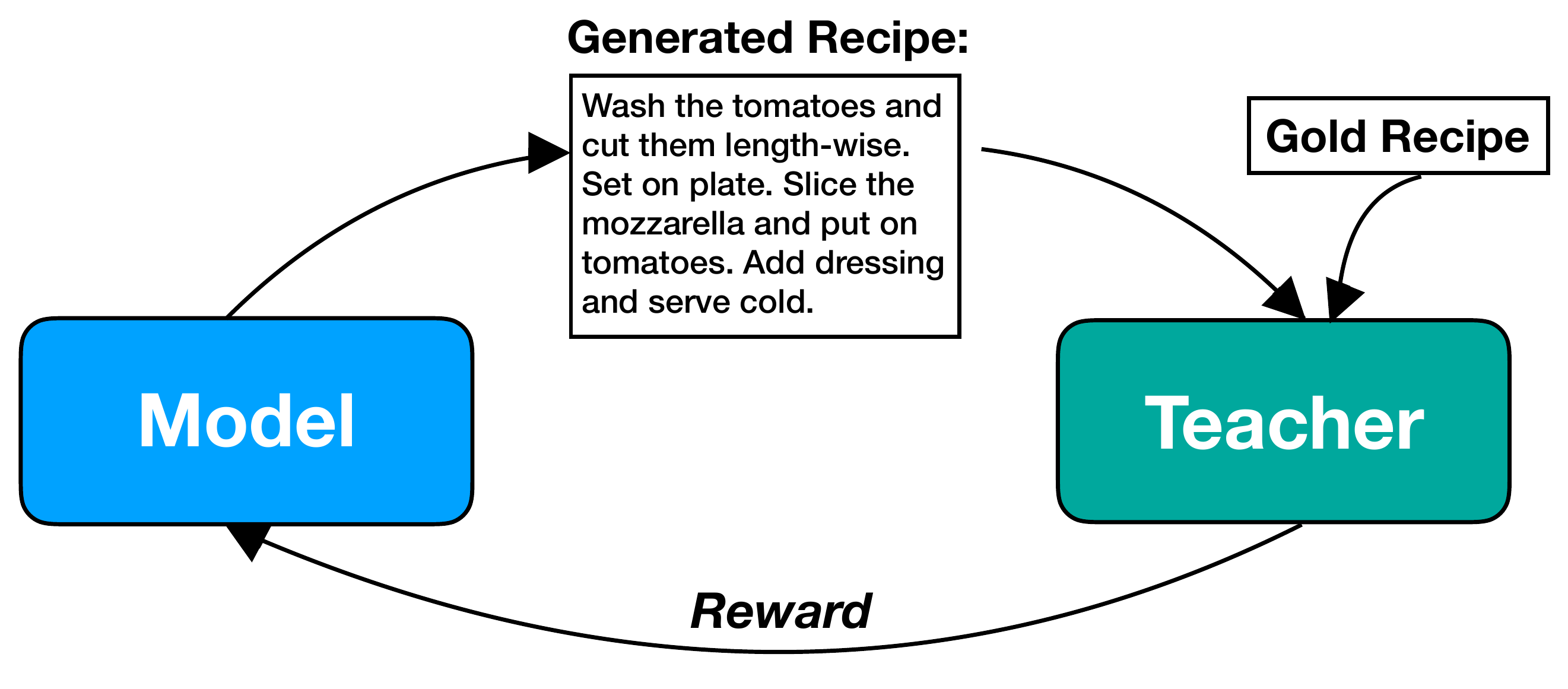}
\caption{The generator is rewarded for imitating the discourse structure of the gold sequence. }
\label{fig:intro}
\end{figure}

Importantly, most automatic measures are based on local $n$-gram patterns, providing only a limited and myopic perspective of overall text quality. As a result, while models trained to directly optimize these measures can yield improvements on the same measures, they may not lead to better quality in terms of overall coherence or discourse structure. Indeed, recent studies have reported cases where commonly used measures do not align well with desired aspects of generation quality \cite{scic,dialoguerl}.  


The challenge, however, is to define a global score that can measure the complex aspects of text quality beyond local $n$-gram patterns. In this paper, we investigate \emph{learning} neural rewards and their use in a reinforcement learning regime with a specific focus on learning more discourse-aware and coherent text generation. Our approach shares the  spirit of the work of \citet{adem}, where neural scores were learned to approximate human judgments of dialogue quality. The key difference is that our rewards can be fully automatically constructed without requiring human judgments and can be trained in an unsupervised manner. 


More specifically, we propose a neural reward learning scheme that is trained to capture cross-sentence ordering structure as a means to approximate the desired discourse structure in documents. 
The learned \emph{teacher} computes rewards for the underlying text generator (see Figure~\ref{fig:intro}), which is trained using self-critical reinforcement learning \mbox{\cite{scic}}. We also present a new method for distributing sentence-level rewards for more accurate credit assignment. 



We test our approach on the task of generating cooking recipes, and evaluate using automatic overlap metrics that measure discourse structure. We also provide human judgments that yield comprehensive insights into the model behavior induced by the learned neural rewards. Empirical results demonstrate that a generator trained with the discourse-aware rewards produces text that is more coherent and less repetitive than models trained with cross-entropy or reinforcement learning with other commonly used scores.

%% file: reward.tex
\section{Neural Teachers}
\label{sec:reward}

Recent work in image captioning \cite{scic}, machine translation \cite{googlemt}, and summarization \cite{rlsummsocher} has investigated using policy gradient methods to fine-tune neural generation models using automatic measures such as CIDEr as the reward. 
However, because most existing automatic measures focus on local $n$-gram patterns, fine-tuning on those measures 
may yield deteriorated text despite increased automatic scores, especially for tasks that require long coherent generation (\S\ref{ssec:exps:overlap}). 

%
Since writing out a scoring term that quantifies the quality of discourse coherence is an open research question, we take inspiration from previous research that \emph{learns} the overall ordering structure of a document as an approximation of the discourse structure \cite{coherence,coherence2,Barzilay2004CatchingTD,coherence3}, 
%
and propose two neural teachers that 
can learn to score an ordered sequence of sentences.
The scores from these neural teachers are then used to formulate rewards (\S\ref{ssec:rl:rewards}) that guide coherent long text generation systems in a policy gradient reinforcement learning setup. Notably, the neural teachers are trained offline on gold sequences in an unsupervised manner prior to training the generator. 
They are not trained jointly with the generator and their parameters are fixed during policy learning.  

\subsection{Notation} We define a document of $n$ sentences as $S = \{s_0, ..., s_n\}$ where each sentence $s_j$ has $L_j$ words.

\begin{figure}[t]
\includegraphics[width=0.48\textwidth]{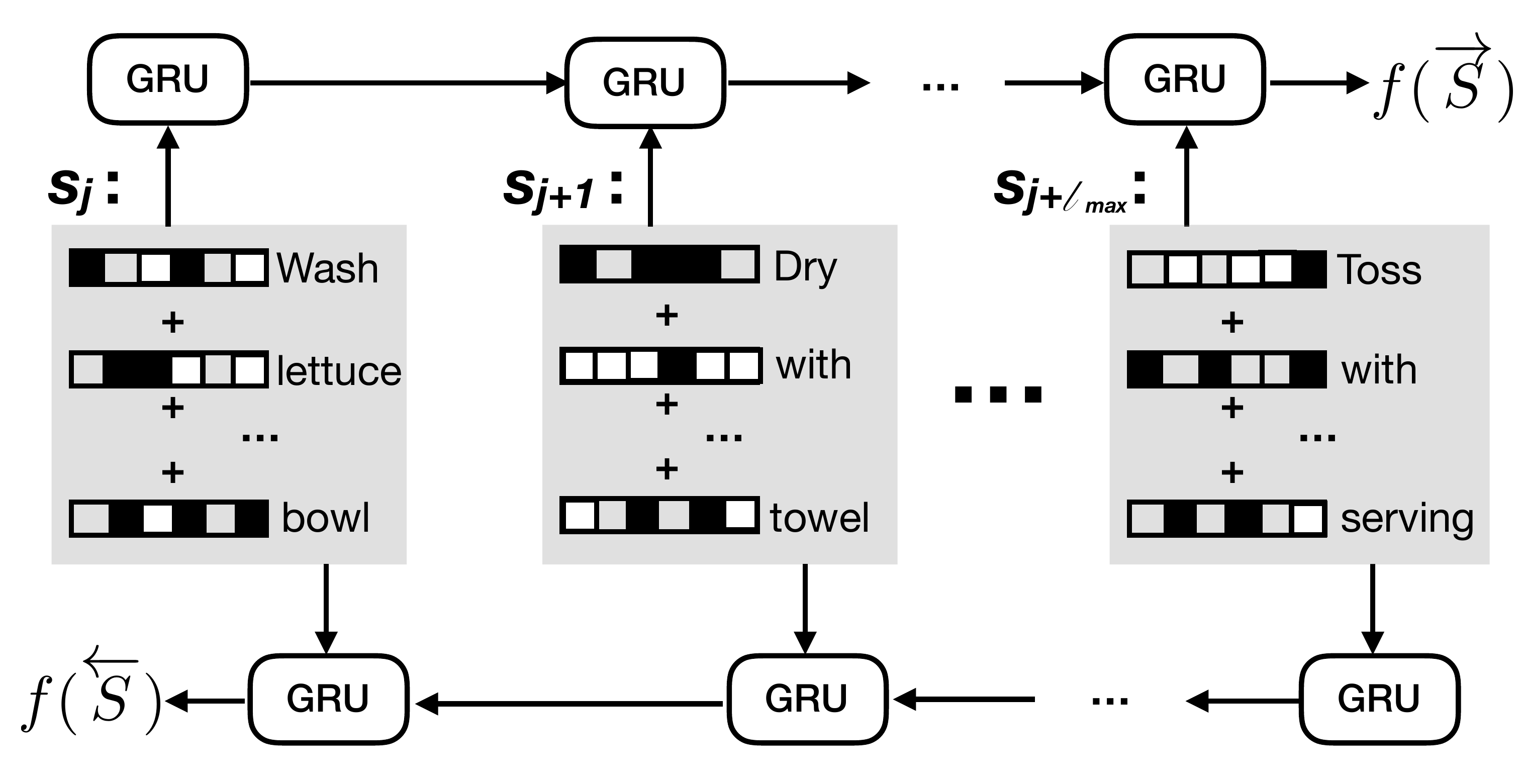}
\caption{The teacher encodes the sentences of the document in the forward and reverse order. }
\label{fig:reward}
\end{figure}

\subsection{Absolute Order Teacher}
\label{ssec:reward:abs}

The first teacher explored is motivated by work on deep semantic similarity models \cite{dssm}, which approximated the similarity between queries and documents in information retrieval tasks. We extend this approach to modeling temporal patterns by training a sentence encoder to minimize the similarity between a sequence encoded in its forward order, and the same sequence encoded in the reverse order (see Figure~\ref{fig:reward}). 

To focus the teacher on discourse structure, we design the encoder to capture \emph{sentence order}, instead of \emph{word order}. Words in each sentence $s_j$ are encoded using a bag of words:

\begin{equation}
\mathbf{s}_j = \sum_{i=1}^{L_j} x_{ij} 
\end{equation}

\noindent where $x_{ij}$ is a word embedding and $\mathbf{s}_j$ is a sentence embedding. Each $\mathbf{s}_j$ is passed to a gated recurrent unit (GRU) and the final output of the hidden unit is used as the representation for the full document:

\begin{align}
h_j = \text{GRU}&(\mathbf{s}_j, h_{j-1}) \\
f(S) &= h_n
\label{eq:aoencode}
\end{align}

\noindent where $f(S)$ is the representation of the sentences of the document and $h_n$ is the final output vector of the GRU. To capture properties of temporal coherence among document sentences, the teacher is trained to minimize $\mathcal{L}_{abs}$, the cosine similarity between the sentence embedding from reading the sentences in the forward order,  $\overrightarrow{S}$ and from reading the sentences in the reverse order,  $\overleftarrow{S}$:

\begin{equation}
\mathcal{L}_{abs} = \frac{ \langle f(\overrightarrow{S}),  f(\overleftarrow{S})\rangle} {\norm{f(\overrightarrow{S})} \norm{f(\overleftarrow{S})}}
\label{eq:aoteacher}
\end{equation}

\noindent Intuitively, by parametrizing only relations between \emph{sentences} (with the GRU layer) and not those between \emph{words}, 
the teacher only captures sentence ordering properties. 
When training the neural generator (\S\ref{sec:rl}), we use this learned teacher to generate a reward that judges the generated sequence's ordering similarity to the gold sequence. 

\subsection{Relative Order Teacher}
\label{ssec:reward:rel}
While the absolute ordering teacher evaluates the temporal coherence of the entire generation, we may want our teacher to be able to judge finer-grained patterns between sentences. In recipes, for example, where sentences correspond to process steps, the teacher should capture implicit script knowledge \cite{schank1975scripts}  among groups of sentences. Consequently, the teacher should reward sentences individually for how they fit with surrounding sentences.

In many current approaches for using policy gradient methods to optimize a model with respect to a global score, 
each sentence receives the same reward. 
This framework assumes each sentence is equally responsible for the reward gathered by the full sequence, allowing potentially appropriate subsequences to be incorrectly penalized. 
We design the relative order teacher to address this issue. 

The relative order teacher is trained in the same way as the absolute order model. A bag of words embedding is computed for each sentence in the gold sequence. Subsequences of the gold document that have $\ell$ sentences are selected where $\ell \in (\ell_{min}$, $\ell_{max}$). For a subsequence beginning at sentence $j$, the model computes:

\begin{align}
f(S_{j:j+\ell}) = \text{GRU}(\mathbf{s}_{j+\ell}, h_{j+ \ell -1})
\label{eq:roencode}
\end{align}

\noindent where $f(S_{j:j+\ell})$ is the encoded representation of sentences $\{s_j,...s_{j+\ell}\}$ and $h_{j-1}$ would be initialized as a vector of zeros. The relative ordering teacher is trained to minimize $\mathcal{L}_{rel}$, the cosine similarity between gold orders of subsequences: 

\begin{equation}
\mathcal{L}_{rel} = \frac{ \langle f(\overrightarrow{S}_{j:j+\ell}),  f(\overleftarrow{S}_{j:j+\ell})\rangle} {\norm{f(\overrightarrow{S}_{j:j+\ell})} \norm{f(\overleftarrow{S}_{j:j+\ell})}}
\label{eq:roteacher}
\end{equation}

\noindent where the arrow above $S$ signifies the order in which the sentences are processed. The relative ordering teacher learns to identify local sentence patterns among ordered sentences, thereby learning how to reward sequences that are temporally coherent.


%% file: agent.tex
\section{Generator Architecture}
\label{sec:gen}
\noindent In the task of recipe generation, the model is given a title of a recipe such as \emph{``Cheese Sandwich"} and a list of ingredients (e.g., cheese, bread, etc.) and must generate the full multi-sentence recipe text. Similar to data to document generation tasks, the model must generate a full long-form text from sparse input signal, filling in missing information on its own \cite{data2docgen}. 

\subsection{Notation}
\label{ssec:gen:not}

Using the same notation as \citet{checklist}, we are given a set of recipe title words $\{g_1, ..., g_n\}$ (e.g., \{ ``cheese", ``sandwich" \}) and a list of ingredients $E = \{\mathbf{i}_1,...,\mathbf{i}_{\vert E \vert}\}$ where each $\mathbf{i}$ can be a single- or multi-word ingredient phrase (e.g., ``onions" or ``onions, chopped"). In the following paragraphs, all $W$ variables are projections matrices and all $b$ variables are bias vectors.

\subsection{Encoder}
\label{ssec:gen:enc}
We use a modification of the baseline encoder of \citet{checklist}. First, the title words are encoded as a bag of embeddings, $\mathbf{g}$. Second, each ingredient phrase $\mathbf{i}$ is encoded as a bag of embeddings vector, $\mathbf{e_i}$. The ingredient embeddings are inputs to a bidirectional gated recurrent unit, which yields an output vector $\mathbf{e}$. The final encoder output is the concatenation of these two representations, $\mathbf{h}^e = [\mathbf{g}, \mathbf{e}]$. 


\subsection{Decoder}
\label{ssec:gen:dec}
The decoder is a separate gated recurrent unit that receives $\mathbf{h}^e$ from the encoder to initialize its hidden state $\mathbf{h}^d_0$ and must generate a full recipe word by word. At each time step, the model receives an input token embedding, $x_t$, as well as the output from the encoder $\mathbf{h}^e$:

\begin{equation}
a_t = \sigma (W_1 h^d_{t - 1} + W_2 x_t + b_1) 
\end{equation}
\vspace{-5mm}
\begin{equation}
\mathbf{z}_t = a_t \mathbf{h}^e
\end{equation}
\vspace{-5mm}
\begin{equation}
 \tilde x_t = [x_t,  \mathbf{z}_t]
\end{equation}

\noindent where $\tilde x_t$ is the input to the recurrent unit at every time step. The recipe generator is pretrained to minimize the negative loglikelihood of predicting the next token in the recipe:

\begin{equation}
L_{mle} = - \sum_{t=1}^{T} \log P(x_{t} | x_0, ..., x_{t-1}, \mathbf{h}^e) \label{eq:mle}
\end{equation}

\noindent where $\mathbf{h}^e$ is the encoded representation of the title and ingredients from Section~\ref{ssec:gen:enc} and $T$ is the number of words in the gold recipe. 

%% file: reinforce.tex
\section{Policy Learning}
\label{sec:rl}


\begin{figure}[t]
\centering
\includegraphics[width=\linewidth]{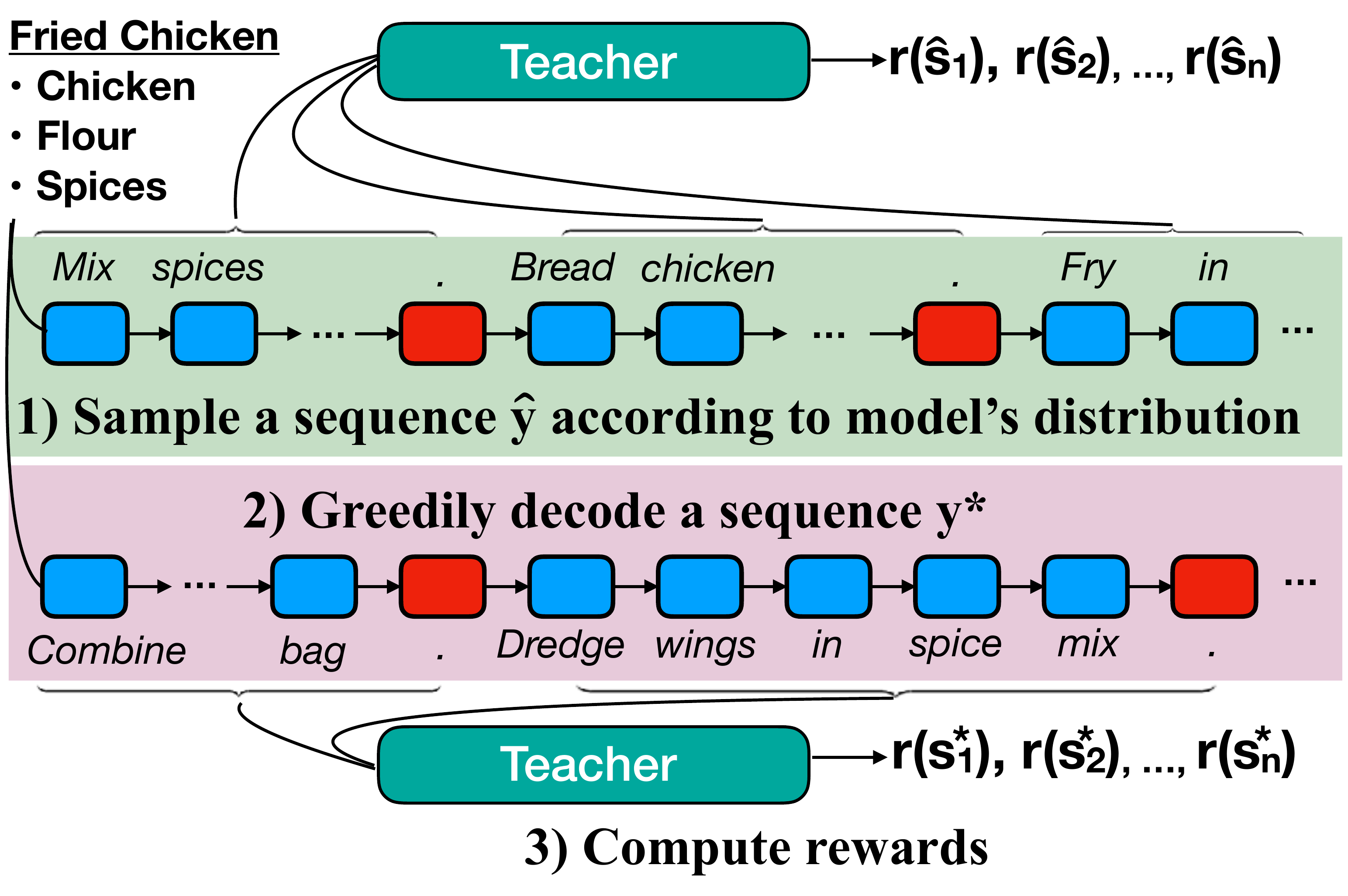}
\caption{The model generates a recipe by sampling from its output vocabulary distribution and greedily decodes a baseline recipe. The generated sentences are passed to the teacher, which yields a reward for each sentence in each recipe. }
\label{fig:selfcritique}
\end{figure}

Training a recipe generation model using maximum likelihood estimation produces generations that are locally coherent, but lack understanding of domain knowledge. By using a teacher that rewards the model for capturing cooking recipe discourse semantics, the model learns a policy that produces generations that better model the underlying recipe process. We learn a policy using the self-critical approach of \citet{scic}.

\subsection{Self-critical sequence training}
\label{ssec:rl:sc}

In self-critical sequence training, outlined in Figure~\ref{fig:selfcritique}, the model learns by being rewarded for sampling sequences that receive more reward than a greedily decoded sequence. For each training example, a sequence ${\hat y}$ is generated by sampling from the model's distribution $P(\hat y_t \vert \hat y_0, ..., \hat y_{t-1}, \mathbf{h}^e)$ at each time step $t$. 
Once the sequence is generated, the teacher produces a reward $r({\hat y_t})$ for each token in the sequence. A second sequence $y^*$ is generated by argmax decoding from $P(y_t^* \vert y_0^*, ..., y_{t-1}^*, \mathbf{h}^e)$ at each time step $t$. The model is trained to minimize:
\begin{equation}
    \mathcal{L}_{rl} = - \sum_{t=1}^{T} (r(\hat y_t) - r(y^*_t)) \log P(\hat y_t \vert \hat y_0, ..., \hat y_{t-1}, \mathbf{h}^e)
    \label{eq:rl}
\end{equation}

\noindent where $r(y^*_t)$ is the reward produced by the teacher for tokens of the greedily decoded sequence. Because $r(y^*)$ can be viewed as a \emph{baseline} reward that sampled sequences should receive more than, the model learns to generate sequences that receive more reward from the teacher than the best sequence that can be greedily decoded from the current policy. This approach allows the model to \emph{explore} sequences that yield higher reward than the current best policy.

\subsection{Rewards} 
\label{ssec:rl:rewards}

As we decode a sequence $y = \{y_0 ..., y_t \}$, we track a sentence index that is the number of sentence delimiter tokens (e.g., ``.") generated by the model. The model then implicitly decodes a set of generated sentences, $S' = \{ s_0, ..., s_n\}$. These sentences are provided to the teachers defined in Section~\ref{sec:reward}, which compute a score for the generated sequence. We explain the procedure for producing a token reward $r(y_t)$ from these scores below. 

\paragraph{Absolute Order}

Once a sequence has been generated, the absolute order teacher computes a reward for $y$ in the following way: 
\begin{equation}
    r_{abs}(y) = \frac{\langle f(S'), f(\overrightarrow{S}) \rangle } {\norm{f(S')}\norm{f(\overrightarrow{S})}} - \frac{ \langle f(S'), f(\overleftarrow{S}) \rangle } {\norm{f({S'})}\norm{f(\overleftarrow{S})}}
    \label{eq:aoreward}
\end{equation}

\noindent where $\overrightarrow{S}$ is the forward-ordered corresponding gold sequence and $\overleftarrow{S}$ is the reverse-ordered gold sequence. Both terms in the reward computation are variations of the loss function on which the absolute order teacher was trained (Equation~\eqref{eq:aoteacher}). This reward compares the generated sequence to both sentence orders of the gold sequence, and rewards generations that are more similar to the forward order of the gold sequence. Because the cosine similarity terms in Equation~\eqref{eq:aoreward} are bounded in $[-1,1]$, the model receives additional reward for generating sequences that are different from the reverse-ordered gold sequence.
\paragraph{Relative Order} Similarly, the relative order reward is generated by the relative order teacher (\S\ref{ssec:reward:rel}), which evaluates subsequences of sentences, rather than the whole sequence. For a sentence $s_j$, the reward is computed as:
\begin{equation}
\begin{aligned}
    r_{rel}( s_j) = \frac{1}{L} \sum_{\ell = \ell_{min}}^{\ell_{max}} \Bigg( \frac{\langle f(S'_{j-\ell:j}), f(\overrightarrow{S}_{j-\ell:j})\rangle}{\norm{f( S'_{j-\ell:j})} \norm{f(\overrightarrow{S}_{j-\ell:j})}} \\
     - \frac{\langle f(S'_{j-\ell:j}), f(\overleftarrow{S}_{j-\ell:j})\rangle} {\norm{f( S'_{j-\ell:j})}\norm{f(\overleftarrow{S}_{j-\ell:j})}}\Bigg)
     \label{eq:roreward}
\end{aligned}
\end{equation}

\noindent where $\ell_{min}$ and $\ell_{max}$ define the window of sentences to include in the computation of the reward. Similar to the absolute order teacher, the relative order teacher produces scores bounded in $[-1,1]$, giving the model additional reward for generating sequences that are different from the reverse-ordered gold subsequences.

\paragraph{Credit Assignment}

When rewarding tokens with the absolute ordering teacher, each generated token receives the same sequence-level reward from the absolute order teacher: \begin{equation}
    r(y_t) = r_{abs}(y)
    \label{eq:rwao}
\end{equation} The relative order teacher, meanwhile, computes rewards for sentences based on their imitation of nearby sentences in the gold recipe. Rather than combining all rewards from the teacher to compute a full sequence reward, sentences should only be rewarded for their own quality. Each token in a sentence corresponds to a position in the full sequence. When relative order rewards are computed by the teacher, the correct sentence reward is indexed for each token. 
Consequently, when training with a relative order teacher, words only receive rewards for the sentences they belong to: 
\vspace{-2mm}
\begin{equation}
    r(y_t) = \sum_{j=1}^{\vert S \vert} \mathbb{1}(y_t \in \hat s_j) r_{rel}(\hat s_j)
    \label{eq:rwro}
\end{equation}
\noindent where $\vert S \vert$ is the number of sentences in the generated recipe, and $\mathbb{1}$ is an indicator variable identifying word $y_t$ belonging to sentence $s_j$.

\subsection{Mixed Training} 
\label{ssec:rl:mixed}
As the model learns parameters to optimize the amount of reward it receives from the teacher, it is not explicity encouraged to produce fluent generations. The model quickly learns to generate simple sequences that \emph{exploit} the teacher for high rewards despite being incoherent recipes (e.g., Figure~\ref{fig:repeat}). Consequently, it is possible that generated sequences are no longer readable \cite{scvc,rlsummsocher}.

 
\begin{figure}[h]
    \small
    \textbf{Title: } Chili Grits \\
    \textbf{Ingredients: } boiling water, butter, shredded cheddar cheese, jalapenos, eggs, chicken cream of soup, salt  \\
    \textbf{Generated Recipe: } \colorbox{pink}{Here .} \\
    \caption{Recipe generated from a self-critical model with no mixed training}
    \label{fig:repeat}
\end{figure}
 
\noindent To remedy this effect, the model optimizes a mixed objective that balances learning the discourse-focused policy while maintaining the generator's language model: 

\begin{equation}
    \mathcal{L}_{mix} = \gamma \mathcal{L}_{rl} + (1-\gamma) \mathcal{L}_{mle}
    \label{eq:mixed}
\end{equation} where $L_{mle}$ is the objective from Equation~\eqref{eq:mle}, $L_{rl}$ is the objective from either Equation~\eqref{eq:rl}, and $\gamma$ is a hyperparameter in [0, 1].

%% file: setup.tex
\section{Experimental Setup}
\label{sec:setup}

\subsection{Datasets}
\label{ssec:setup:data}
We use the Now You're Cooking dataset with the same training/test/development splits from \citet{checklist}. For training, we use 109567 recipes with 1000 recipes set aside for both development and test. 

\subsection{Training}
\label{ssec:setup:train}

\paragraph{Teacher Models} The teachers are trained before the recipe generator and their parameters are fixed during generation. We tune hyperparameters on the development set. To train the relative order teacher, we sample 20 subsequences from each recipe of $\ell_{min} = 3$ to $\ell_{max} = 6$ sentences. Additional details are provided in Appendix~\ref{appendix:reward}.

\paragraph{Recipe Generator} We pretrain a recipe generator using a variant of the encoder-decoder baseline from \citet{checklist}. Comprehensive hyperparameter details can be found in Appendix~\ref{appendix:pretrain}.

\begin{table*}[t]
\resizebox{\linewidth}{!}{
    \centering
    \begin{tabular}{r || r | r | r || r | r | r || r | r | r }
         Model & BLEU-1 & BLEU-4 & R-L & AB1 & AB4 & AR-L & SCB1 & SCB4 & SCR-L \\
         \hline\hline
         Cross-entropy (MLE) & 26.86 & 4.74 & 28.86 & 31.23 & 4.83 & 28.51 & 51.92 & 26.35 & 50.21 \\
         \hline
        BLEU-4 \cite{scic} & 7.75 & 1.38 & 13.93 & 5.69 & 0.84 & 10.37 & 10.76 & 5.05 & 20.87 \\
        CIDEr \cite{scic} & 12.67 & 1.90 & 21.20 & 14.61 & 1.79 & 21.70 & 26.07 & 12.30 & 41.65 \\
         ROUGE-L \cite{rlsummsocher} & 29.00 & 4.86 & 29.10 & 33.49 & 4.73 & 28.11 & 56.86 & 27.83 & 51.26\\ \hline
        BLEU-1 ($\gamma = 0.97$) & \textbf{31.16} & \textbf{5.60} & 29.53 & 32.28 & 5.09 & 29.34 & 52.63 & 25.43 & 51.58 \\
        BLEU-4 ($\gamma = 0.99$) & 30.56 & 5.42 & 29.16 & 32.53 & 4.99 & 28.99 & 53.48 & 26.35 & 51.02 \\
        CIDEr ($\gamma = 0.97$) & 29.60 & 5.10 & 28.79 & 33.93 & 4.81 & 28.41 & 57.00 & 27.55 & 50.57 \\
        ROUGE-L ($\gamma = 0.97$) & 26.88 & 4.66 & 29.49 & 31.85 & 5.01 & 29.25 & 53.84 & 26.77 & 51.88 \\
        \hline
         Absolute Ordering (AO)  & 23.70 & 4.25 & 28.43 & 28.22 & 4.44 & 27.88 & 47.93 & 24.47 & 50.15 \\
        Relative Ordering (RO) & 27.75 & 4.88 & 29.60 & 34.37 & \textbf{5.60} & \textbf{29.36} & 58.31 & 29.14 & \textbf{53.08} \\
         \hline         
         Relative Ordering + BLEU-4 & 29.58 & 5.26 & \textbf{29.78} & \textbf{35.13} & 5.55 & 29.33 & \textbf{59.13} & \textbf{29.19} & 52.46 \\
    \end{tabular}}
    \caption{Evaluation results for generated sequences by models and baselines. We \textbf{bold} the top performing result. The second to fourth columns list word-level scores. Columns AB1, AB4, and AR-L list action-level scores (\S\ref{ssec:exps:overlap}). Columns SCB1, SCB4, and SCR-L list state change level scores (\S\ref{ssec:exps:overlap}).}
    \label{tab:quant}
\end{table*}

\paragraph{Policy Learning} 
We train a different model for three different teacher-provided rewards: absolute ordering (AO), relative ordering (RO) and a joint reward of relative ordering and BLEU-4 (RO + B4), where the full-sequence BLEU-4 reward and the sentence-level relative ordering reward are summed at each time step. The best model for the absolute and relative ordering rewards are the ones that receive the highest average reward on the development set. The best model for the mixed reward was chosen as the one that achieved the highest average geometric mean of BLEU-4 reward and average relative ordering reward for each generated sequence $y$ in the development set:

\begin{equation}
\bar r = \frac{r_{b4}(y)}{T}\sum_{t=1}^T r_{RO}(y_t)
\end{equation}

\noindent where $r_{b4}$ is the BLEU-4 score of the whole generated sequence, and $r_{RO}$ is computed using Equation~\eqref{eq:rwro}. Our best models use $\gamma = 0.97$ when training with the mixed objective from Equation~\eqref{eq:mixed}.



\subsection{Baselines}
\label{ssec:setup:baselines}
As baselines, we report results for a model trained only with cross-entropy loss (MLE) and for re-implemented versions of models from \citet{scic} and \citet{rlsummsocher}. These baselines achieved state of the art results in image captioning and document summarization tasks. We found, however, that their high $\gamma$ (1 and 0.9984, respectively) led to low fluency, resulting in reduced performance on word-level scores. To control for this effect, we trained additional versions of each baseline with different values for $\gamma$ and report the best performing configurations (see Table~\ref{tab:quant}). 

%% file: experiments.tex
\section{Results}
\label{sec:experiments}

\input{exps/overlap}
\input{exps/human}
\input{exps/insights}

%% file: exps/overlap.tex
\subsection{Overlap Metrics}
\label{ssec:exps:overlap}

\paragraph{Scores} We compute the example-level BLEU-1, BLEU-4, and ROUGE-L (R-L) scores for all recipes in the test set.
A generated recipe, however, must be coherent at both the \emph{word-level}, linking words and phrases sensibly, and the \emph{world-level}, describing events that are grounded in real-world actions.
Because $n$-gram scores do not evaluate if a generated recipe models this latent process, we also report these scores on the \emph{action} and \emph{state change} sequence described in the recipe. 
These words depict a \emph{simulated} world where actions are taken and state changes are induced. A generated recipe should follow the sequence of actions taken in the gold recipe, and induce the same state changes as those in the gold recipe. 

We use the state change lexicon from \citet{npn} to map recipe words to ordered sequences of actions and state changes. Each entry in the lexicon contains an action in the cooking domain as well as the state changes that result from that action in the set of \{\textsc{location, composition, cookedness, temperature, shape, cleanliness}\}. 

Action sequences are formed by mapping lemmas of words in generated sequences to entries in the lexicon. We compare these event sequences to the gold event sequences using the same scores as for words -- BLEU-1, BLEU-4, and ROUGE-L. Intuitively, these scores can be seen as evaluating the following: whether the generated recipe depicts the same actions (AB1), subsequences of consecutive actions (AB4), and full action sequence (AR-L) as the gold recipe. 

State change sequences are more coarse-grained than action sequences, and are formed by mapping actions to their state changes in the lexicon from \citet{npn}. These scores evaluate whether the generated recipe implies the same induced state changes (SCB1), subsequences of consecutive state changes (SCB4), and global state change order (SCR-L) as the gold recipe.
\paragraph{Results}
Our results in Table~\ref{tab:quant} show that models optimized on word overlap metrics achieve the greatest improvements for those scores.  Optimizing scores such as BLEU-1 encourages the model to output words and phrases that overlap often with reference sequences, but that may not describe main events in the recipe process. 

When examining models trained using a neural teacher, we see that the model optimized with the absolute ordering reward performs worse than most baselines for every word-level score. The relative ordering model, however, raises every word-level score above the cross-entropy baseline, indicating the importance of fine-grained credit assignment at the sentence-level. The model trained with mixed rewards from the teacher and BLEU-4 achieves even higher scores, showing the benefits of training with diverse rewards. 

When evaluating these metrics for the action and state change sequence, the models trained with feedback from the relative ordering teacher show large improvement over the baselines, indicating that the models exhibit more understanding of the latent process underlying the task. While optimizing word-level scores teaches the generator to output common sequences of words, the relative ordering reward teaches the model to focus on learning co-occurrences between recipe events.

%% file: exps/human.tex
\subsection{Human Evaluation}
\label{ssec:exps:human}

\begin{table}[t]
    \centering
    \begin{tabular}{r || r r r }
          & MLE & RO + B4 & Tie \\
         \hline
         Fluency  & 0.330 & \textbf{0.447} & 0.223  \\
         Ingredient Use  & 0.350 & \textbf{0.440} & 0.210 \\
         Title Completion & 0.347 & \textbf{0.430} & 0.223 \\
         Action Order & 0.377 & \textbf{0.453} & 0.170 \\
         \hline
         & BLEU-1 & RO + B4 & Tie \\
         \hline
         Fluency  & \textbf{0.387} & 0.373 & 0.240  \\
         Ingredient Use  & 0.327 & \textbf{0.363} & 0.310 \\
         Title Completion & 0.353 & \textbf{0.377} & 0.270 \\
         Action Order & \textbf{0.410} & 0.403 & 0.187 \\
    \end{tabular}
    \caption{Human evaluation measuring proportion of winners. Upper table compares MLE baseline with RO + B4 model. Lower table compares BLEU-1 baseline with RO + B4 model.}
    \label{tab:human_all}
\end{table}

We perform a human evaluation on 100 recipes sampled from the test set to evaluate our model on four aspects of recipe quality: fluency, ingredient use, title completion, and action ordering. For each example, three judges from Amazon Mechanical Turk are shown a pair of recipes, each generated by a different model and asked to select the recipe that is better according to the criteria above. For ingredient use, judges select the recipe that uses more of the ingredients correctly. For title completion, we ask judges to select the recipe that best completes the dish described in the recipe title. Finally, for action ordering, judges choose the recipe that better links subtasks in the recipes. 
\paragraph{Models} We use the Relative Ordering + BLEU-4 model (RO + B4) and compared to two baselines, the cross-entropy model (MLE), and the BLEU-1 model, which achieved the best scores on several word-level metrics (\S\ref{ssec:exps:overlap}).
\paragraph{Results} We report results in Table~\ref{tab:human_all}. Our model outperforms the cross-entropy baseline, consistently being preferred on aggregate for every question. Workers preferred the BLEU-1 baseline for the fluency and action order questions, while preferring recipes generated by the teacher-trained model for the ingredient use and title ordering questions. Upon further analysis, we see that the strength of the BLEU-1 model depends on the length of the original reference sequence. In Table~\ref{tab:human_100}, we show evaluation scores for recipes where the gold recipe was longer than 100 words. Our model's performance rises compared to the BLEU-1 model for every question, showing that modeling discourse structure as learned reward improves global coherence in long text.  


\begin{table}[t]
    \centering
    \begin{tabular}{r || r r r }
          & MLE & RO + B4 & Tie \\
         \hline
         Fluency  & 0.317 & \textbf{0.425} & 0.258  \\
         Ingredient Use  & 0.342 & \textbf{0.458} & 0.200 \\
         Title Completion & 0.358 & \textbf{0.450} & 0.192 \\
         Action Order & 0.367 & \textbf{0.483} & 0.150 \\
         \hline
         & BLEU-1 & RO + B4 & Tie \\
         \hline
         Fluency  & \textbf{0.391} & 0.383 & 0.225  \\
         Ingredient Use  & 0.267 & \textbf{0.392} & 0.342 \\
         Title Completion & 0.325 & \textbf{0.418} & 0.258 \\
         Action Order & 0.433 & \textbf{0.442} & 0.125 \\
    \end{tabular}
    \caption{Proportion of winners for \textbf{long} generated recipes. Upper table compares MLE baseline with RO + B4 model. Lower table compares BLEU-1 baseline with mixed RO + B4 model.}
    \label{tab:human_100}
\end{table}

%% file: exps/insights.tex
\subsection{Insights}
\label{ssec:insights}

\input{exps/examples}

\begin{figure}[t!]
\includegraphics[width=0.5\textwidth]{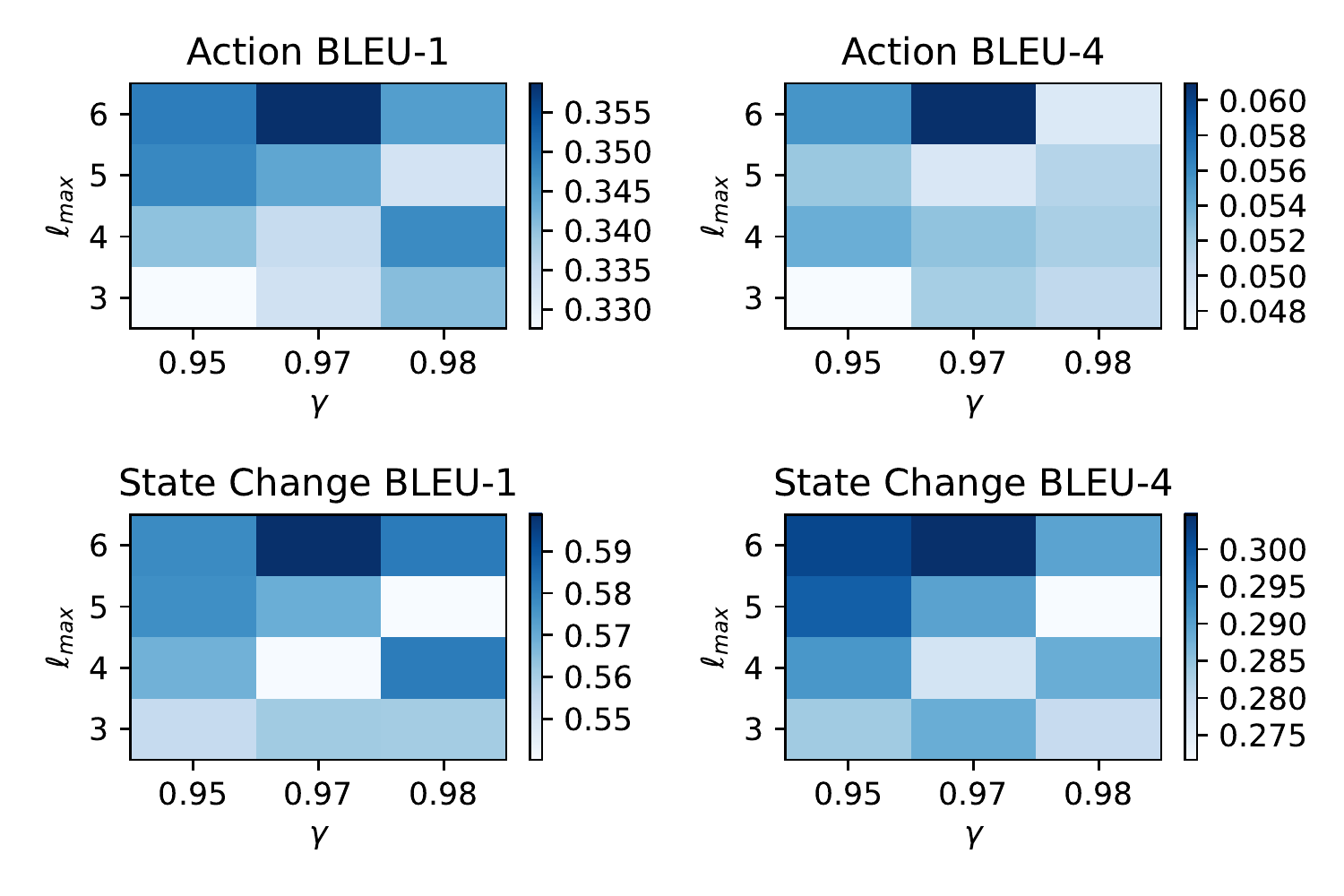}
\caption{Action and State Change BLEU Metrics for different initializations of $\ell_{max}$ and $\gamma$}
\label{fig:heatmap}
\end{figure}

\paragraph{Qualitative Analysis} In Table~\ref{table:genex}, we see the effect that the neural teacher has on the recipe generator. The teacher rewards behavior that more closely imitates the actions in the gold recipe. In the first example, the generator learns to complete the actions of placing the mixture into the a \emph{greased casserole} and then \emph{baking} it, which the MLE model misses. The teacher also discourages repetitive phrases, as they provide no increase in reward during training. One weakness of our teacher models, however, is that they encourage common temporal patterns, such as in the third example in Table~\ref{table:genex}, where the generator mentions \emph{baking the pie}. The model recognizes pies are generally supposed to be baked, even if it is not appropriate for that particular recipe.
\paragraph{Teacher Feedback Frequency} We design the reward functions in Eq.~\ref{eq:aoreward} and Eq.~\ref{eq:roreward} to require two passes through the teacher, one comparing the generated sequence to the forward gold sequence, and one comparing it to the reverse gold sequence.  
With no teacher comparison to the reverse-ordered sequence, the generator learns to \emph{exploit} the teacher for reward with very simple sequences such as ``Serve." and ``Here's direction." When comparing with both orders, however, this effect is dampened, hinting at the importance of ensembling feedback from multiple sources for robust reward production. Another solution to this effect was mixing policy learning and maximum likelihood learning (Eq.~\ref{eq:mixed}) as the underlying language model of the generator did not deteriorate.
\paragraph{Impact of $\ell_{max}$ and $\gamma$}
Two hyperparameters to tune when training with teacher models are the mixed loss coefficient $\gamma$, which balances MLE learning with policy learning, and [$\ell_{min}$, $\ell_{max}$], the number of sentences to consider when computing the relative order reward. We fix $\ell_{min} = 3$, and vary $\ell_{max} \in [3, 6]$ and $\gamma \in \{0.95, 0.97, 0.98\}$. Figure~\ref{fig:heatmap} shows the importance of tuning $\gamma$. A low $\gamma$ will not allow the teacher to guide the model's learning, while a high $\gamma$ causes the language model to deteriorate. Interestingly, a higher $\ell_{max}$ leads to better performance on global coherence scores, implying that relative order rewards conditioned on more sentences allow the model to learn longer-range context co-occurrences.

%% file: exps/examples.tex
\begin{table*}[t]
\small
\begin{center}
\begin{tabular}{| r | l |}
\hline
\textbf{Title} & Royale Casserole\\
\textbf{Ings} & Broccoli, chopped; butter; flour; salt; black pepper; nutmeg; milk; cheese shredded swiss; lemon juice; \\
& cooked cubed turkey; mushrooms; grated parmesan cheese; biscuits, refrigerated \\
\hline
\textbf{MLE} & Cook broccoli according to package directions. Drain. Melt butter in skillet. Add flour, stirring constantly. \\
& Cook, stirring constantly, until thickened. Add milk, salt, pepper, and nutmeg. Cook over low heat, stirring \\
& constantly, until thickened. Add cheese, salt, pepper and pimento. Cook and stir till thickened. Add cheese, \\
& stirring until melted. Add \fbox{\textbf{\color{blue} broccoli and mushrooms}}. Cook over low heat for 5 minutes. \colorbox{pink}{Add cheese. Cook} \\
& \colorbox{pink}{and stir until cheese melts. Add cheese. Cook and stir until cheese melts. Add cheese. Cook and stir until} \\
& \colorbox{pink}{cheese melts. Add cheese and stir until melted. Serve over hot noodles.} \\
\hline
\textbf{RO} & Melt butter in saucepan. Add flour and cook, stirring constantly, until thickened. Add milk, salt, pepper, \\
& and nutmeg. Cook, stirring constantly, until thickened. Add cheese, salt and pepper. Cook over low heat, \\ 
& stirring constantly, until mixture is thickened and smooth. Remove from heat. Stir in cheese. Spoon into \\
& \fbox{\textbf{\color{blue}greased casserole}}. Top with cheese. \fbox{\textbf{\color{blue}Bake in 350 f oven for 30 minutes.}} Serves 6. \\ 
\hline
\textbf{Gold} & Preheat oven to 375. Melt butter in saucepan. Blend in flour, salt, pepper, and nutmeg; cook 1-2 minutes. \\
& Gradually add milk; cook, stirring, until slightly thickened. Stir in frozen vegetables. Remove from heat; stir \\
& in cheese until melted. Add lemon juice, turkey, mushrooms and broccoli. Pour mixture into a lightly greased 
\\ & baking dish; sprinkle with parmesan and top with biscuits. Bake 20 minutes, or until biscuits are golden brown. \\
\hline 
\hline
\textbf{Title} & Wine Sauerkraut\\
\textbf{Ings} & butter; dry white wine; 16 oz drained sauerkraut \\
\hline
\textbf{MLE} & In a saucepan, combine \colorbox{pink}{soup}, wine, and butter. Heat, uncovered, in microwave oven 2 minutes or until \\
& mixture is heated through. \\
\hline
\textbf{RO} & Melt butter in skillet. \fbox{\textbf{\color{blue} Add sauerkraut}} and wine; heat to boiling. Reduce heat; cover and simmer 15 minutes. \\
& Add wine and heat to boiling; reduce heat. Cover and cook 10 minutes or until mixture is tender. Serve on rice.\\
\hline
\textbf{Gold} & Melt butter in 10-inch skillet over low heat; add sauerkraut. Cover and cook, stirring occasionally, 30 \\
& minutes; add wine. Heat to boiling; reduce heat. Cover and simmer until liquid is absorbed, about 45 \\
& minutes. Follow directions except simmer until liquid is absorbed, about 1 hour. \\
\hline\hline
\textbf{Title} & Strawberry Glazed Pie\\
\textbf{Ings} & fresh strawberries; granulated sugar; salt; cornstarch; lemon juice; baked, cooled pie shell \\
\hline
\textbf{MLE} & Combine sugar, cornstarch and salt in top of a double boiler. Cook and stir over medium heat until mixture \\ 
& thickens and boils. Boil and stir 1 minute. Remove from heat and stir in lemon juice. Cool. Add strawberries \\ 
& \colorbox{pink}{and lemon juice}. Pour into baked pie shell. Top with whipped cream. Serve immediately. \\
\hline
\textbf{RO} & Combine sugar, cornstarch, and salt in a saucepan; cook over medium heat, stirring constantly, until \\
& mixture boils and thickens. Remove from heat; stir in lemon juice. Cool. Add strawberries; mix well. Pour \\
& into baked pie shell. \colorbox{pink}{Bake in preheated 325-degree oven for 10 minutes}. Cool. Garnish with whipped cream. \\
\hline
\textbf{Gold} & Wash, drain thoroughly, and hull strawberries. Arrange about 3 cups of whole berries over bottom of baked \\
& pastry shell. Crush remaining berries in a saucepan. In a bowl, mix sugar, salt and cornstarch; stir into crushed \\
& berries. Heat slowly, stirring constantly, until mixture comes to a boil and thickens. Remove from heat and stir \\
& in lemon juice. Cool, then spoon over berries in pie shell chill until glaze is set. Garnish with whipped cream. \\
\hline
\end{tabular}
\end{center}
\caption{Example recipe generations from our model and comparative baselines. \fbox{\textbf{\color{blue}Boxed}} spans indicate recipe events missed by another model's generation. \colorbox{pink}{Red} spans indicate superfluous events. The \textbf{Ings} row lists the ingredients (separated by semicolons) provided to make the dish in the title.}
\label{table:genex}
\end{table*}

%% file: related.tex
\section{Related Work}
\label{sec:related}

The field of neural text generation has received considerable attention in tasks such as image captioning \cite{NIC,ImgAttn}, summarization \cite{deepsummrush,summpointernet}, machine translation \cite{bahdanau2014neural}, and recipe generation \cite{checklist}. While these works have focused on developing new neural architectures that introduce structural biases for easier \emph{learning}, our work uses a simple architecture and focuses on improving the optimization of the learner (i.e., better \emph{teaching}).

The importance of better teaching for RNN generators was outlined in \citet{scheduledsampling}, which showed that \emph{exposure} bias from a misaligned train and test setup limited the capabilities of sequence-to-sequence models. This limitation had been addressed in previous work by augmenting training data with examples generated by pretrained models to make models robust to their own errors \cite{searn,dagger}.

More recent work on training RNNs for generation has used sequence scores such as ROUGE \cite{rlsummsocher}, CIDEr \cite{scic,scvc}, BLEU \cite{mixer} and mixtures of them \cite{spider} as a global reward to train a policy with the REINFORCE algorithm \cite{reinforce}. In contrast, our work uses a neural teacher to reward a model for capturing discourse semantics.

Most similar to our work is work on using neural and embedding rewards to improve dialogue \cite{dialoguerl}, image captioning \cite{rlemb}, simplification \cite{rlsimp}, and paraphrase generation \cite{rlparateacher}. While these works use single-sentence similarity rewards for short generation tasks, our work designs teachers to reward long-range ordering patterns.

Finally, our teachers can be seen as rewarding generators that approximate script patterns in recipes. Previous work in learning script knowledge \cite{schank1975scripts} has focused on extracting scripts from long texts \cite{chambers2009unsupervised,pichottascripts}, with some of that work focusing on recipes \cite{actiongraph,mori2014flow,mori2012machine}. 
Our teachers implicitly learn this script knowledge and reward recipe generators for exhibiting it.

%% file: conclusion.tex
\section{Conclusion}
\label{sec:conclusions}

We introduce the \emph{absolute ordering} and \emph{relative ordering} teachers, two neural networks that score a sequence's adherence to discourse structure in long text. The teachers are used to compute rewards for a self-critical reinforcement learning framework, allowing a recipe generator to be rewarded for capturing temporal semantics of the cooking domain. Empirical results demonstrate that our teacher-trained generator better models the latent event sequences of cooking recipes, and a human evaluation shows that this improvement is mainly due to maintaining semantic coherence in longer recipes.

\section*{Acknowledgments}
We thank Chloe Kiddon for helpful discussions in the early stages of this work. This research was supported in part by NSF (IIS-1524371), DARPA under the CwC program through the ARO (W911NF-15-1-0543) and Samsung Research.

%% file: appendix.tex
\appendix

\section{Hyperparameters}

\subsection{Data}

Each recipe is batched based on the number of tokens and number of ingredients it has. We use a minibatch size of 32.

\subsection{Teachers}
\label{appendix:reward}
The hidden size of the reward generator is 100, the word embeddings have dimensionality 100. We use dropout with a rate of 0.3 between the bag of words layers and the recurrent layers. 

\subsection{Pretrained Recipe Generator}
\label{appendix:pretrain}
We use a hidden size of 256 for the encoder and 256 for the decoder. We initialize three different sets of embeddings for the recipe titles, ingredient lists, and text, each of size 256. All models are trained with a dropout rate of 0.3 and are single-layer. We use a temperature coefficient of $\beta = 2$ to make the output word distribution more peaky \cite{checklist}, allowing for more controlled exploration during self-critical learning. We use scheduled sampling with a linear decay schedule of 5\% every 5 epochs up to a max of 50\%. We use a learning rate of $\eta = 0.0003$ and train with the Adam optimizer.

\subsection{Policy Learning}
\label{appendix:policy}
We use the same model hyperparameters as during pretraining, but re-initialize the Adam optimizer, use $\eta = 3 \times 10^{-5}$ as the learning rate, and do not train with scheduled sampling. 

\section{Baseline Selection}
For each baseline we trained, we report the score of the $\gamma$ setting that achieved the highest score for the metric on which it was trained. For example, for baselines trained with ROUGE-L reward, we report the results for the model trained with the value of $\gamma$ that scored the highest ROUGE-L score on the development set. For the models trained with the CIDEr reward, we select the model with value of $\gamma$ that achieved the highest CIDEr score on the development set. We do the same for models trained with BLEU-1 and BLEU-4 rewards. The values of $\gamma$ yielding the best performance on the development set were $0.97$ for the BLEU-1, ROUGE-L, and CIDEr-trained models, and $0.99$ for the BLEU-4 trained baseline. For each baseline, the best model is chosen by selecting the checkpoint that achieves the highest reward (or lowest loss for the MLE model) for the metric it was trained on.

%% file: main.bbl
\begin{thebibliography}{}
\expandafter\ifx\csname natexlab\endcsname\relax\def\natexlab#1{#1}\fi

\bibitem[{Bahdanau et~al.(2015)Bahdanau, Cho, and Bengio}]{bahdanau2014neural}
Dzmitry Bahdanau, Kyunghyun Cho, and Yoshua Bengio. 2015.
\newblock Neural machine translation by jointly learning to align and
  translate.
\newblock In {\em Proceedings of the 3rd International Conference for Learning
  Representations\/}.

\bibitem[{Barzilay and Lapata(2005)}]{coherence}
Regina Barzilay and Mirella Lapata. 2005.
\newblock Modeling local coherence: An entity-based approach.
\newblock In {\em Proceedings of the 43rd Annual Meeting of the Association for
  Computational Linguistics\/}.

\bibitem[{Barzilay and Lapata(2008)}]{coherence2}
Regina Barzilay and Mirella Lapata. 2008.
\newblock Modeling local coherence: An entity-based approach.
\newblock {\em Computational Linguistics\/} 34(1).

\bibitem[{Barzilay and Lee(2004)}]{Barzilay2004CatchingTD}
Regina Barzilay and Lillian Lee. 2004.
\newblock Catching the drift: Probabilistic content models, with applications
  to generation and summarization.
\newblock In {\em HLT-NAACL\/}.

\bibitem[{Bengio et~al.(2015)Bengio, Vinyals, Jaitly, and
  Shazeer}]{scheduledsampling}
Samy Bengio, Oriol Vinyals, Navdeep Jaitly, and Noam Shazeer. 2015.
\newblock Scheduled sampling for sequence prediction with recurrent neural
  networks.
\newblock In {\em Advances in Neural Information Processing Systems\/}.

\bibitem[{Bosselut et~al.(2018)Bosselut, Levy, Holtzman, Ennis, Fox, and
  Choi}]{npn}
Antoine Bosselut, Omer Levy, Ari Holtzman, Corin Ennis, Dieter Fox, and Yejin
  Choi. 2018.
\newblock Simulating action dynamics with neural process networks.
\newblock {\em Proceedings of the 6th International Conference for Learning
  Representations\/} .

\bibitem[{Chambers and Jurafsky(2009)}]{chambers2009unsupervised}
Nathanael Chambers and Dan Jurafsky. 2009.
\newblock Unsupervised learning of narrative schemas and their participants.
\newblock In {\em Proceedings of the Joint Conference of the 47th Annual
  Meeting of the ACL and the 4th International Joint Conference on Natural
  Language Processing of the AFNLP: Volume 2-Volume 2\/}. Association for
  Computational Linguistics.

\bibitem[{Cho et~al.(2014)Cho, van Merrienboer, Gulcehre, Bougares, Schwenk,
  and Bengio}]{cho2014learning}
Kyunghyun Cho, Bart van Merrienboer, Caglar Gulcehre, Fethi Bougares, Holger
  Schwenk, and Yoshua Bengio. 2014.
\newblock Learning phrase representations using rnn encoder-decoder for
  statistical machine translation.
\newblock In {\em Proceedings of the 2014 Conference on Empirical Methods in
  Natural Language Processing\/}.

\bibitem[{{Daum\'e III} et~al.(2009){Daum\'e III}, Langford, and Marcu}]{searn}
Hal {Daum\'e III}, John Langford, and Daniel Marcu. 2009.
\newblock Search-based structured prediction .

\bibitem[{Hochreiter and Schmidhuber(1997)}]{lstm}
Sepp Hochreiter and J{\"u}rgen Schmidhuber. 1997.
\newblock Long short-term memory.
\newblock {\em Neural Computation\/} 9(8).

\bibitem[{Huang et~al.(2013)Huang, He, Gao, Deng, Acero, and Heck}]{dssm}
Po-Sen Huang, Xiaodong He, Jianfeng Gao, Li~Deng, Alex Acero, and Larry Heck.
  2013.
\newblock Learning deep structured semantic models for web search using
  clickthrough data.
\newblock In {\em Proceedings of the 22nd ACM International Conference on
  Information \& Knowledge Management\/}. ACM.

\bibitem[{Kiddon et~al.(2015)Kiddon, Ponnuraj, Zettlemoyer, and
  Choi}]{actiongraph}
Chlo{\'e} Kiddon, Ganesa~Thandavam Ponnuraj, Luke Zettlemoyer, and Yejin Choi.
  2015.
\newblock Mise en place: Unsupervised interpretation of instructional recipes.
\newblock In {\em Proceedings of the 2015 Conference on Empirical Methods in
  Natural Language Processing\/}.

\bibitem[{Kiddon et~al.(2016)Kiddon, Zettlemoyer, and Choi}]{checklist}
Chlo{\'e} Kiddon, Luke Zettlemoyer, and Yejin Choi. 2016.
\newblock Globally coherent text generation with neural checklist models.
\newblock In {\em Proceedings of the 2016 Conference on Empirical Methods in
  Natural Language Processing\/}.

\bibitem[{Li and Hovy(2014)}]{coherence3}
Jiwei Li and Eduard~H Hovy. 2014.
\newblock A model of coherence based on distributed sentence representation.
\newblock In {\em Proceedings of the 2014 Conference on Empirical Methods in
  Natural Language Processing\/}.

\bibitem[{Li et~al.(2016)Li, Monroe, Ritter, Galley, Gao, and
  Jurafsky}]{dialoguerl}
Jiwei Li, Will Monroe, Alan Ritter, Michel Galley, Jianfeng Gao, and Dan
  Jurafsky. 2016.
\newblock Deep reinforcement learning for dialogue generation.
\newblock In {\em Proceedings of the 2016 Conference on Empirical Methods in
  Natural Language Processing\/}.

\bibitem[{Li et~al.(2017)Li, Jiang, Shang, and Li}]{rlparateacher}
Zichao Li, Xin Jiang, Lifeng Shang, and Hang Li. 2017.
\newblock Paraphrase generation with deep reinforcement learning.
\newblock {\em arXiv preprint arXiv:1711.00279\/} .

\bibitem[{Lin(2004)}]{rouge}
Chin-Yew Lin. 2004.
\newblock {ROUGE}: a package for automatic evaluation of summaries.
\newblock In {\em Text summarization branches out: Proceedings of the ACL-04
  workshop\/}. Barcelona, Spain, volume~8.

\bibitem[{Liu et~al.(2017)Liu, Zhu, Ye, Guadarrama, and Murphy}]{spider}
Siqi Liu, Zhenhai Zhu, Ning Ye, Sergio Guadarrama, and Kevin Murphy. 2017.
\newblock Improved image captioning via policy gradient optimization of spider.
\newblock {\em Proceedings of the 2017 IEEE International Conference on
  Computer Vision\/} .

\bibitem[{Lowe et~al.(2017)Lowe, Noseworthy, Serban, Angelard-Gontier, Bengio,
  and Pineau}]{adem}
Ryan Lowe, Michael Noseworthy, Iulian Serban, Nicolas Angelard-Gontier, Yoshua
  Bengio, and Joelle Pineau. 2017.
\newblock Towards an automatic turing test: Learning to evaluate dialogue
  responses.
\newblock In {\em Proceedings of the 55th Annual Meeting of the Association for
  Computational Linguistics\/}.

\bibitem[{Mori et~al.(2014)Mori, Maeta, Yamakata, and Sasada}]{mori2014flow}
Shinsuke Mori, Hirokuni Maeta, Yoko Yamakata, and Tetsuro Sasada. 2014.
\newblock Flow graph corpus from recipe texts.
\newblock In {\em Proceedings of the Ninth International Conference on Language
  Resources and Evaluation\/}.

\bibitem[{Mori et~al.(2012)Mori, Sasada, Yamakata, and
  Yoshino}]{mori2012machine}
Shinsuke Mori, Tetsuro Sasada, Yoko Yamakata, and Koichiro Yoshino. 2012.
\newblock A machine learning approach to recipe text processing.
\newblock In {\em Proceedings of the 1st Cooking with Computer Workshop\/}.

\bibitem[{Papineni et~al.(2002)Papineni, Roukos, Ward, and Zhu}]{bleu}
Kishore Papineni, Salim Roukos, Todd Ward, and Wei-Jing Zhu. 2002.
\newblock {BLEU}: a method for automatic evaluation of machine translation.
\newblock In {\em Proceedings of the 40th Annual Meeting of the Association for
  Computational Linguistics\/}. Association for Computational Linguistics.

\bibitem[{Pasunuru and Bansal(2017)}]{scvc}
Ramakanth Pasunuru and Mohit Bansal. 2017.
\newblock Reinforced video captioning with entailment rewards.
\newblock In {\em Proceedings of the 2017 Conference on Empirical Methods in
  Natural Language Processing\/}.

\bibitem[{Paulus et~al.(2018)Paulus, Xiong, and Socher}]{rlsummsocher}
Romain Paulus, Caiming Xiong, and Richard Socher. 2018.
\newblock A deep reinforced model for abstractive summarization.
\newblock In {\em Proceedings of the 6th International Conference for Learning
  Representations\/}.

\bibitem[{Pichotta and Mooney(2016)}]{pichottascripts}
Karl Pichotta and Raymond~J. Mooney. 2016.
\newblock Using sentence-level lstm language models for script inference.
\newblock In {\em Proceedings of the 54th Annual Meeting of the Association for
  Computational Linguistics\/}.

\bibitem[{Ranzato et~al.(2015)Ranzato, Chopra, Auli, and Zaremba}]{mixer}
Marc'Aurelio Ranzato, Sumit Chopra, Michael Auli, and Wojciech Zaremba. 2015.
\newblock Sequence level training with recurrent neural networks.
\newblock In {\em Proceedings of the 4th International Conference for Learning
  Representations\/}.

\bibitem[{Ren et~al.(2017)Ren, Wang, Zhang, Lv, and Li}]{rlemb}
Zhou Ren, Xiaoyu Wang, Ning Zhang, Xutao Lv, and Li-Jia Li. 2017.
\newblock Deep reinforcement learning-based image captioning with embedding
  reward.
\newblock {\em Proceedings of the 2017 IEEE Conference on Computer Vision and
  Pattern Recognition\/} .

\bibitem[{Rennie et~al.(2017)Rennie, Marcheret, Mroueh, Ross, and Goel}]{scic}
Steven~J. Rennie, Etienne Marcheret, Youssef Mroueh, Jarret Ross, and Vaibhava
  Goel. 2017.
\newblock Self-critical sequence training for image captioning.
\newblock {\em Proceedings of the 2017 IEEE Conference on Computer Vision and
  Pattern Recognition\/} .

\bibitem[{Ross et~al.(2011)Ross, Gordon, and Bagnell}]{dagger}
St{\'e}phane Ross, Geoffrey~J Gordon, and Drew Bagnell. 2011.
\newblock A reduction of imitation learning and structured prediction to
  no-regret online learning.
\newblock In {\em International Conference on Artificial Intelligence and
  Statistics\/}.

\bibitem[{Rush et~al.(2015)Rush, Chopra, and Weston}]{deepsummrush}
Alexander~M. Rush, Sumit Chopra, and Jason Weston. 2015.
\newblock A neural attention model for abstractive sentence summarization.
\newblock In {\em Proceedings of the 2015 Conference on Empirical Methods in
  Natural Language Processing\/}.

\bibitem[{Schank and Abelson(1975)}]{schank1975scripts}
Roger~C Schank and Robert~P Abelson. 1975.
\newblock {\em Scripts, plans, and knowledge\/}.
\newblock Yale University.

\bibitem[{See et~al.(2017)See, Liu, and Manning}]{summpointernet}
Abigale See, Peter~J. Liu, and Christopher Manning. 2017.
\newblock Gettothepoint: Summarization with pointer-generatornetworks.
\newblock In {\em Proceedings of the 55th Annual Meeting of the Association for
  Computational Linguistics\/}.

\bibitem[{Vedantam et~al.(2015)Vedantam, Lawrence~Zitnick, and Parikh}]{cider}
Ramakrishna Vedantam, C~Lawrence~Zitnick, and Devi Parikh. 2015.
\newblock {CIDEr}: Consensus-based image description evaluation.
\newblock In {\em Proceedings of the 2015 IEEE Conference on Computer Vision
  and Pattern Recognition\/}.

\bibitem[{Vinyals et~al.(2015)Vinyals, Toshev, Bengio, and Erhan}]{NIC}
Oriol Vinyals, Alexander Toshev, Samy Bengio, and Dumitru Erhan. 2015.
\newblock Show and tell: A neural image caption generator.
\newblock In {\em Proceedings of the 2015 IEEE Conference on Computer Cision
  and Pattern Recognition\/}.

\bibitem[{Williams(1992)}]{reinforce}
Ronald~J Williams. 1992.
\newblock Simple statistical gradient-following algorithms for connectionist
  reinforcement learning.
\newblock {\em Machine learning\/} 8(3-4).

\bibitem[{Wiseman et~al.(2017)Wiseman, Shieber, and Rush}]{data2docgen}
Sam Wiseman, Stuart~M. Shieber, and Alexander~M. Rush. 2017.
\newblock Challenges in data-to-document generation.
\newblock In {\em Proceedings of the 2017 Conference on Empirical Methods in
  Natural Language Processing\/}.

\bibitem[{Wu et~al.(2016)Wu, Schuster, Chen, Le, Norouzi, Macherey, Krikun,
  Cao, Gao, Macherey et~al.}]{googlemt}
Yonghui Wu, Mike Schuster, Zhifeng Chen, Quoc~V Le, Mohammad Norouzi, Wolfgang
  Macherey, Maxim Krikun, Yuan Cao, Qin Gao, Klaus Macherey, et~al. 2016.
\newblock Google's neural machine translation system: Bridging the gap between
  human and machine translation.
\newblock {\em arXiv preprint arXiv:1609.08144\/} .

\bibitem[{Xu et~al.(2015)Xu, Ba, Kiros, Cho, Courville, Salakhutdinov, Zemel,
  and Bengio}]{ImgAttn}
Kelvin Xu, Jimmy Ba, Ryan Kiros, Kyunghyun Cho, Aaron~C. Courville, Ruslan
  Salakhutdinov, Richard~S. Zemel, and Yoshua Bengio. 2015.
\newblock Show, attend and tell: Neural image caption generation with visual
  attention.
\newblock In {\em Proceedings of The 32nd International Conference on Machine
  Learning\/}.

\bibitem[{Yang et~al.(2017)Yang, Blunsom, Dyer, and Ling}]{reflm}
Zichao Yang, Phil Blunsom, Chris Dyer, and Wang Ling. 2017.
\newblock Reference-aware language models.
\newblock In {\em Proceedings of the 2017 Conference on Empirical Methods in
  Natural Language Processing\/}.

\bibitem[{Zhang and Lapata(2017)}]{rlsimp}
Xingxing Zhang and Mirella Lapata. 2017.
\newblock Sentence simplification with deep reinforcement learning.
\newblock In {\em Proceedings of the 2017 Conference on Empirical Methods in
  Natural Language Processing\/}.

\end{thebibliography}
